\documentclass[runningheads]{llncs}
\usepackage[T1]{fontenc}
\usepackage{graphicx}
\usepackage{hyperref}
\usepackage{color}

\usepackage[autolanguage]{numprint}
\usepackage{booktabs}
\usepackage{microtype}
\usepackage{bbding}

\usepackage{ifthen}
\newboolean{authorv}
\setboolean{authorv}{true}

\ifthenelse{\boolean{authorv}}{
\hypersetup{
    colorlinks=true,
    linkcolor=[RGB]{0 0 0},
    urlcolor=[RGB]{0 0 0},
    citecolor=[RGB]{0 0 0},
    pdfborder={0 0 0},
    frenchlinks=false
}
\usepackage{orcidlink}
\renewcommand{\orcidID}{\orcidlink}
}

\begin{document}

\title{Extracting Information in a Low-resource Setting: Case Study on Bioinformatics Workflows}
\titlerunning{Information Extraction in Bioinformatics Workflows}

\author{Cl\'emence Sebe\inst{1}\orcidID{0000-0003-1988-1875} \and
Sarah Cohen-Boulakia\inst{1}\orcidID{0000-0002-7439-1441} \and
Olivier Ferret\inst{2}\orcidID{0000-0003-0755-2361}
\and
Aur\'elie N\'ev\'eol\inst{1}\orcidID{0000-0002-1846-9144}}
\authorrunning{C. Sebe et al.}
\institute{Universit\'e Paris-Saclay, LISN, CNRS, Orsay, 91400, France \email{firstname.lastname.@lisn.fr} \and
Université Paris-Saclay, CEA, List, F-91120, Palaiseau, France
\\
\email{olivier.ferret@cea.fr}
}
\maketitle 

\begin{abstract} Bioinformatics workflows are essential for complex biological data analyses and are often described in scientific articles with source code in public repositories. Extracting detailed workflow information from articles can improve accessibility and reusability but is hindered by limited annotated corpora.
To address this, we framed the problem as a low-resource extraction task and tested four strategies: 1) creating a tailored annotated corpus, 2) few-shot named-entity recognition (NER) with an autoregressive language model, 3) NER using masked language models with existing and new corpora, and 4) integrating workflow knowledge into NER models.
Using BioToFlow, a new corpus of 52 articles annotated with 16 entities, a SciBERT-based NER model achieved a 70.4 F-measure, comparable to inter-annotator agreement. While knowledge integration improved performance for specific entities, it was less effective across the entire information schema.
Our results demonstrate that high-performance information extraction for bioinformatics workflows is achievable.

\end{abstract}

\section{Introduction}
In data-intensive sciences, such as \textit{bioinformatics}, scientific results are produced using complex data analysis pipelines taking in vast amounts of experimental data. These pipelines consist of numerous steps making calls to bioinformatics tools and require considerable computational time. They can be implemented using scripts (Bash, Python, etc.) that orchestrate tool execution and serve as the glue between data, processes, tools, and the execution environment. However, the development and use of such scripts lead to challenges in the design and execution of the processing chain, as well as maintenance and reproducibility issues. 
In response to these challenges, efforts have been made over the past two decades to provide bioinformaticians with approaches that guide them toward better automation of data analysis pipelines: the \textit{scientific workflow systems} \cite{cohenboulakia}. Two systems 
stand out today in bioinformatics: Nextflow \cite{nextflow} and Snakemake \cite{snakemake}.

Currently, we observe: (1) an increase in scientific articles describing bioinformatics workflows in a \textit{descriptive} form (steps described without being executable) and (2) a rise in bioinformatics workflows available on repositories like GitHub that are in a \textit{programmatic} form (implementation available but the lack of documentation makes them difficult to reuse). 

In Natural Language Processing (NLP), bioinformatics workflows remain underrepresented. To our knowledge, there is no comprehensive corpus encompassing a rich and diverse set of entities describing workflows (bioinformatics tools, data, management systems, etc.). However, 
some corpora focus on specific types of entities, such as those centered on bioinformatics tools \cite{du_softcite_2021,istrate_large_2022}. This lack of resources makes bioinformatics a low-resource context, making it difficult to structure information or analyze texts with the limited annotated data available.

To overcome the low-resource challenges in the field, we consider in this paper several strategies. 
First, decoder-based approaches, such as those using generative models, offer a potential solution, although they may not always outperform other methods in terms of extraction accuracy \cite{chen_systematic_2024,naguib_few-shot_2024}. Next, encoder-based approaches can be explored in two ways: using larger, related corpora to enrich model training, or by developing a small, domain-specific training corpus to focus on the extraction of entities relevant to the field. Finally, another promising approach is the injection of domain-specific knowledge directly into language models. 

From a methodological perspective, we address in this paper several research questions related to low-resource settings. In a specific domain, what resources should be annotated, in what quantity and how should domain experts be involved? Which existing resources can be reused or adapted for the new task? What methods are the most effective for the target task and how can they be combined?
From a more application-oriented perspective, the challenge is to develop a simple and automated approach capable of extracting information about workflows described in the literature, aiming to not only systematically document workflows presented in the literature but also facilitate their registration in workflow repositories. 

In preliminary work~\cite{sebeTALN2024}, we presented a corpus of scientific articles annotated with entities representing bioinformatics workflow information (BioToFlow). We also reported preliminary experiments with Named-Entity Recognition (NER) from the corpus with a specific focus on the memorization vs. generalization abilities of statistical and rule-based methods.
Herein, we provide a brief description of BioToFlow and we present new contributions as follows:
(a) experiments with few-shot NER using an autoregressive language model; (b) NER experiments using a pre-existing corpus and BioToFlow and (c) integration of workflow knowledge into NER models. Our code is publicly available\footnote{\url{https://gitlab.liris.cnrs.fr/sharefair/ner\_workflows}}.

\section{Workflow Information  Representation }
\subsection{Workflow Representation Schema} 

\begin{figure}[ht]
\centering
 \includegraphics[width=0.8\textwidth]{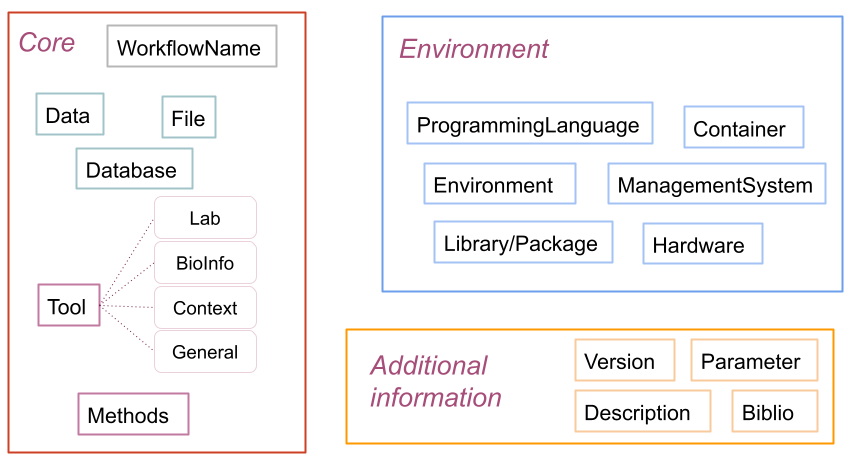}
 \caption{Bioinformatics workflow representation schema.} \label{schemaEntite}
\end{figure}

To systematically describe the bioinformatics workflows and their components, we relied on discussions with workflow experts (notably from the ELIXIR and BOSC International Bioinformatics communities) and the review of several articles describing workflows (e.g., \cite{cohenboulakia,nextflow,kieser_atlas_2020,snakemake,yates_reproducible_2021}. Generally, bioinformatics workflows consist of data analysis steps, each containing a script that may or may not invoke a bioinformatics tool. For a workflow to be executed, it is crucial to maintain a record of its execution environment. 
Figure \ref{schemaEntite} presents the proposed schema, which distinguishes between three main categories of entities. The first category is \textit{core} entities, representing major components of a workflow, such as the name of the workflow described, bioinformatics tools, and data. 
\textit{Tools} can also be further classified based on the context: whether they are explicitly identified as bioinformatics tools (BioInfo), laboratory-related tools (Lab), tools mentioned but not used in the final workflow (Context), or tools referenced in a general sense without specific details (General). One entity is dedicated to methods, i.e. the descriptions of workflow steps, but without mentioning specific tool names (e.g., "mapping DNA sequences against a genome" vs. "BWA").
The data refers to the information processed by the workflow (e.g., "DNA sequence"). Articles exhibit different levels of specificity and can contain mentions of  a database name or the exact filenames that were used. However, in other cases, they remain more general and only refer to broad categories of data. The second category includes \textit{environment} entities, which capture the resources and configurations required for workflow execution, such as the execution environment (e.g., the management system used, the hardware it runs on) or programming languages used in scripts. Finally, the third category addresses more specific details, including versioning information, bibliographic references describing tools or libraries, as well as descriptions and parameters associated.
This additional information can be associated with the entities present in the workflow body or be more closely related to the workflow environment, providing further context about it.

\subsection{Annotation of Workflow Information in Articles: \textit{BioToFlow}}

\subsubsection{Article Selection Criteria} 
Articles describing bioinformatics workflows with a link to workflow code in Github have been targeted using a query on PubMed and PubMedCentral \cite{webPmcMubMed}\footnote{PubMed query: (nextflow[Abstract] OR snakemake[Abstract] OR nextflow[Title] OR snakemake[Title]) AND github[All Fields]}. 
During the annotation phase (February 2024), this query returned 89 and 91 articles describing workflows from Nextflow and Snakemake systems respectively, from which we randomly selected articles for annotation. As of 11/18/2024, the same query now returns 124 and 123 articles, demonstrating a continuous increase in the number of published articles on these workflows. Articles were pre-processed to select the contents of the \textit{Material and Methods} and \textit{Implementation} sections, which commonly contain workflow descriptions that we aim to annotate.

\subsubsection{Creation of an Annotated Corpus} 
To validate the information representation schema described in section 2.1  
and create a resource for evaluating the extraction of such information, we developed an annotated corpus using a collaborative method \cite{fortWILEY2016} with the help of an annotation guide that we designed. Annotations were performed using BRAT (Brat Rapid Annotation Tool \cite{brat}, v1.3 p1), recognized for its qualities including the ability to process nested entities, an existing annotation schema, and pre-annotations \cite{reviewTool}.

\begin{figure}[ht]
\begin{center}
 \includegraphics[width=1.0\textwidth]{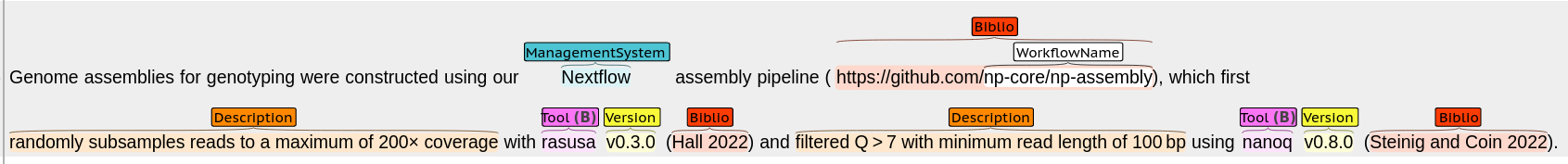}
 \caption{Excerpt from the annotated corpus using the BRAT software.} \label{imageBrat}
\end{center}
\end{figure}

Seven annotators participated in the annotation campaigns, which were conducted in multiple stages: the annotators first worked on common texts, then on different texts. Inter-annotator agreement (IAA) was calculated using the F-measure\footnote{For a detailed review of inter-annotator agreement measures and their usage context, we refer the reader to \cite{artsteinCL2008}. In the case of named entity annotation, \cite{grouin-2011law} demonstrates that the F-measure and $\kappa$ can be considered equivalent.} with BRAT-Eval \cite{verspoor2013annotating}(v0.0.2), a tool that compares two sets of annotations and analyzes divergences. The evaluation of agreement can be strict (agreement between two annotations only if identical text portions are annotated with the same label) or relaxed (agreement if two identical or overlapping text portions are annotated with the same label). The score ranges from 0 (no agreement) to 1 (perfect agreement). 
IAA obtained were all higher than 70\% in relaxed mode. When we look in more detail at the extraction scores for each entity, we found a heterogeneity in the scores showing that some are easier to annotate than others. There are several causes for these differences, including ambiguity in the tagset, criteria for discrimination \cite{fortCOLING2012}, and, finally, the expertise domain of the annotators. Figure \ref{imageBrat} shows entity annotations using our schema on an excerpt from an article describing a Nextflow workflow (PMID 35171290).

Overall, our corpus, named \textit{BioToFlow}\footnote{\url{https://doi.org/10.5281/zenodo.14900544}}, is a small corpus composed of 52 articles (26 articles related to Nextflow workflows and 26 on Snakemake workflows, randomly selected from PubMed while ensuring a balanced representation of both workflow management systems) with a total of \numprint{78 419} tokens (\numprint{27 786} annotated tokens). However, the entities are diversified and distributed according to Table~\ref{repartion_entites}. In our corpus, we allow the presence of nested entities ($\sim$8\% of entities in BioToFlow; an example can be seen in Figure~\ref{imageBrat}, where a WorkflowName entity is nested within a Biblio entity).

\begin{table}[!ht]
\caption{BioToFlow: number of occurrences per entity.} \label{repartion_entites}
\centering
\setlength{\tabcolsep}{6pt}
\begin{tabular}{lclc}
\toprule
\textbf{Entities} & \textbf{Occurrences} & \textbf{Entities} & \textbf{Occurrences}  \\ 
\midrule
 Data         &  \numprint{2434}  &  Version             &   \numprint{454} \\ 
 Tool         &  \numprint{1482}  &  Hardware            &   \numprint{429} \\ 
 Description  &  \numprint{1300}  &  Database            &   \numprint{288} \\ 
 Biblio       &  \numprint{1251}  &  ManagementSystem    &   \numprint{243} \\ 
 Method       &  \numprint{936}    &  Container           &  \numprint{108} \\ 
 WorkflowName &  \numprint{851}    &  ProgrammingLanguage &  \numprint{104} \\ 
 File         &  \numprint{780}    &  LibraryPackage      &  \numprint{101} \\ 
 Parameter    &  \numprint{464}    &  Environment         &  \numprint{83} \\ 
 \bottomrule
\end{tabular}
\end{table}

Using this corpus and its annotations as a starting point, we explored different extraction approaches, starting with decoder-based approaches and then moving on to encoder-based approaches.

\section{Decoder-based Approach for Named Entity Recognition}
Because of the small size of the BioToFlow corpus, we sought to leverage an auto-regressive language model to perform few-shot named entity recognition. Specifically, we used Llama-3-8B-Instruct, which demonstrated the best performance among generative models in a comparative study of masked language models and generative models utilizing prompting techniques \cite{naguib_few-shot_2024}. Following the same methodology as in this study\footnote{The code is available at \url{https://github.com/marconaguib/autoregressive\_ner}}, we conducted a series of few-shot experiments, varying the number of examples  provided (either 5 or 10) and testing different prompt formulations, such as providing or omitting entity definitions and task framing.

We use random sampling to split the BioToFlow corpus:  we use 75\% of the corpus (39 articles) for training and the remaining 25\% (13 articles) for testing.  The results obtained on the BioToFlow test set using the Llama-3-8B-Instruct model with the best hyperparameters are presented in Table \ref{tab_llm_new}.

\nprounddigits{1}
        \begin{table}[!ht]
        \caption{Performance of Llama-3-8B-Instruct model with the best hyperparameters (relaxed mode).}
        \label{tab_llm_new}
        \footnotesize
        \centering
        \setlength{\tabcolsep}{6pt}
        \begin{tabular}{llllllll}
            \toprule
\textbf{Entities}   & \textbf{P} & \textbf{R} & \textbf{F1} & \textbf{Entities}           & \textbf{P} & \textbf{R} & \textbf{F1} \\ \midrule
Biblio      & \numprint{93,10} & \numprint{82,20} & \numprint{87,30}  & LibraryPackage      & \numprint{3,80}  & \numprint{5,60} & \numprint{4,50}  \\
Container   & \numprint{29,20} & \numprint{50,00} & \numprint{36,80}  & ManagementSystem    & \numprint{41,10}  & \numprint{38,30} & \numprint{39,70}  \\
Data        & \numprint{31,20} & \numprint{27,60} & \numprint{29,30}  & Method              & \numprint{15,40}  & \numprint{12,10} & \numprint{13,60}  \\
Database    & \numprint{28,30} & \numprint{24,60} & \numprint{26,30}  & Parameter           & \numprint{11,40}  & \numprint{8,80} & \numprint{9,90}  \\
Description & \numprint{26,20} & \numprint{19,40} & \numprint{22,30}  & ProgrammingLanguage & \numprint{23,50}  & \numprint{14,30} & \numprint{17,80}  \\
Environment & \numprint{60,00} & \numprint{13,00} & \numprint{21,40}  & Tool                & \numprint{54,90}  & \numprint{59,10} & \numprint{56,90}  \\
File        & \numprint{30,70} & \numprint{34,80} & \numprint{32,60}  & Version             & \numprint{67,10}  & \numprint{39,50} & \numprint{49,70}  \\
Hardware    & \numprint{20,50} & \numprint{26,30} & \numprint{23,10}  & WorkflowName        & \numprint{29,10}  & \numprint{37,70} & \numprint{32,80}  \\ \midrule
\textbf{Overall}     & \numprint{40,10} & \numprint{36,70} & \numprint{38,30}  &                     &            &            &                                     
\\ \bottomrule
\end{tabular}
\end{table}
\nprounddigits{0}

\nprounddigits{1}
We obtained an overall performance below 40\% in relaxed mode and \numprint{31.2}\% in strict mode (not shown). 
These results are consistent with findings on other biomedical corpora~\cite{naguib_few-shot_2024}. 
Bioinformatics workflows are complex, and although they are closely related to the medical domain, the entities are often very specific. This leads us to explore alternative methods leveraging encoder-based models.
\nprounddigits{0}

\section{Encoder-based Approach for Named Entity Recognition}

Encoder-based approaches for NER typically require a substantial amount of annotated corpus. In our experiments, we leverage a large corpus with annotations covering a subset of workflow information and a small corpus with comprehensive coverage of workflow information.   

\subsection{Experiments on a Large Corpus: \textit{SoftCite}} \label{section_softcite}

To our knowledge, no corpus fully covers workflows. However, we identified corpora with annotations partially aligned with our schema, such as those focusing on datasets \cite{pan_dmdd_2023} or bioinformatics tools. We focused on bioinformatics tools, as identifying them offers crucial insights into a workflow's functionality. We can cite the CZ Software Mentions dataset \cite{istrate_large_2022}, which includes software mentions extracted using a trained SciBERT model from diverse sources like PubMed Central and the Chan Zuckerberg Initiative. Another relevant corpus is SoftCite \cite{du_softcite_2021}, a gold-standard dataset where all articles were manually annotated. SoftCite comprises \numprint{4971} articles from the fields of biomedicine and economics. We chose SoftCite due to its manual annotation process and, to assess its relevance, we selected a subset of \numprint{1159} articles specifically related to biology.

Since SoftCite does not contain the same entity types as BioToFlow, our first step was to identify possible overlaps between the two entity schemas and perform a mapping to convert SoftCite entity names to those used in BioToFlow. 
Table \ref{tab_correspondance_ent_soft_bio} represents the correspondence between entities in SoftCite (left column) and those in BioToFlow (right column). Some entities in the SoftCite corpus, \textit{publisher\_person}, \textit{figure}, \textit{table}, and \textit{formula}, have no equivalent in BioToFlow. Similarly, certain entities in BioToFlow, including \textit{data}, \textit{container}, \textit{hardware}, and \textit{description}, are not present in SoftCite. After the conversion, the number of entities and occurrences are presented in Table~\ref{tab_ent_softcite}.

\begin{table}[!ht]
\caption{Entity correspondence between SoftCite and BioToFlow.} \label{tab_correspondance_ent_soft_bio}
\footnotesize
\centering
\begin{tabular}{ll@{\hskip 1cm}l}
\toprule
\multicolumn{2}{c}{\textbf{SoftCite}} & \textbf{BioToFlow} \\
\midrule
software  && Tool\\
  & environment& Tool\\
  & url        & Biblio          \\
  & component  & LibraryPackage  \\
  & implicit   & Tool\_general   \\
    publisher && Biblio          \\
  & environment& Environment     \\
    bibr      && Biblio          \\
    version   && Version         \\
    url       && Biblio          \\
    language  && ProgrammingLanguage \\
    \bottomrule
\end{tabular}
\end{table}

\begin{table}[!ht]
\caption{SoftCite: Number of occurrences per entity.} \label{tab_ent_softcite}
\footnotesize
\centering
\setlength{\tabcolsep}{6pt}
\begin{tabular}{llll}
\toprule
\textbf{Entities} & \textbf{Occurrences} & \textbf{Entities} & \textbf{Occurrences} \\
\midrule
Tool & \numprint{4142} & Biblio & \numprint{3946} \\
Version & \numprint{1349} & LibraryPackage & \numprint{65} \\
ProgrammingLanguage & \numprint{61} \\
    \bottomrule
\end{tabular}
\end{table}

For the extraction of our target entities, we chose a biLSTM-CRF neural model (combination of a bidirectional LSTM (for sequence processing) with CRF (for prediction)) \cite{wajsburt_extraction_2021}, implemented by the NLStruct Python library package\footnote{\url{https://github.com/percevalw/NLStruct}} (v0.2.0), which performs multiclass classification. All the experiments with NLStruct were run with the hyperparameters provided in Appendix \ref{nlstruct_h}. This library offers the dual advantage of supporting the detection of nested entities, which is essential for certain entity types in our dataset, and accepting input files in the BRAT format. Additionally, it has demonstrated strong performance in the biomedical domain, which closely aligns with our field. More specifically, we used this model with the SciBERT language model \cite{beltagy_scibert_2019}, more specifically trained on scientific articles. 
We conventionally chose to randomly divide the articles into two main sets: the first set, 75\% of the corpus (927 articles), was used to create the training set, while the remaining 25\% (232 articles) constituted the test set. Within the training set, two-thirds of the articles (649 articles) were used for the actual training of the models, and one-third (278 articles) for their validation. A distinctive aspect of our approach is that we constructed five splits through random sampling to minimize dependency on a specific division. For each split, five versions of the entity recognition model were produced using different random seeds, ensuring more robust and reproducible results.

\nprounddigits{1}
\begin{table}[!ht]
\caption{Average performance metrics on five splits and five random seeds (with standard deviation) of percentage scores obtained using SciBERT and NLStruct in relaxed mode on SoftCite corpora.} \label{tab_softcite}
\footnotesize
\centering
\setlength{\tabcolsep}{6pt}
\begin{tabular}{lllll}
\toprule
 &  \multicolumn{3}{c}{\textbf{Softcite}}       \\ \cmidrule(lr){2-4}
    & \multicolumn{1}{c}{\textbf{P}} & \multicolumn{1}{c}{\textbf{R}} & \multicolumn{1}{c}{\textbf{F1}} \\
    \midrule
    Biblio          & \numprint{90,64}{\scriptsize $\pm$\numprint{1,44}} & \numprint{98,95}{\scriptsize $\pm$\numprint{0,35}}  & \numprint{94,61}{\scriptsize $\pm$\numprint{0,88}}                    \\
    Environment       & \numprint{0,00}{\scriptsize $\pm$\numprint{0,00}}   & \numprint{0,00}{\scriptsize $\pm$\numprint{0,00}}  & \numprint{0,00}{\scriptsize $\pm$\numprint{0,00}}                     \\
    LibraryPackage   & \numprint{2,45}{\scriptsize $\pm$\numprint{2,41}}   & \numprint{14,27}{\scriptsize $\pm$\numprint{10,47}}  & \numprint{4,06}{\scriptsize $\pm$\numprint{3,88}}                    \\
    ProgLanguage    & \numprint{48,86}{\scriptsize $\pm$\numprint{5,31}}  & \numprint{88,83}{\scriptsize $\pm$\numprint{4,76}} & \numprint{62,77}{\scriptsize $\pm$\numprint{4,79}}                    \\
    Tool            & \numprint{73,40}{\scriptsize $\pm$\numprint{1,94}}  & \numprint{51,04}{\scriptsize $\pm$\numprint{0,58}}  & \numprint{60,20}{\scriptsize $\pm$\numprint{0,99}}                    \\
    Version         & \numprint{49,81}{\scriptsize $\pm$\numprint{2,21}}  & \numprint{93,54}{\scriptsize $\pm$\numprint{0,61}} & \numprint{64,97}{\scriptsize $\pm$\numprint{1,88}}                 \\
    \textbf{Overall-focused}         & \numprint{73,14}{\scriptsize $\pm$\numprint{1,11}}  & \numprint{70,71}{\scriptsize $\pm$\numprint{0,60}}  & \numprint{71,90}{\scriptsize $\pm$\numprint{0,75}} \\
    \bottomrule                   
    \end{tabular}
    \end{table}
\nprounddigits{0}

We test the results on 13 test articles from BioToFlow to see if the entities of the SoftCite corpus could be used to train and extract a part of the BioToFlow entities. 
The performance on a subset of entities (Table \ref{tab_softcite}) using this corpus is significantly better than those achieved with a decoder-based approach.

\subsection{Experiments on a Small Corpus:  \textit{BioToFlow}} \label{sec:ner_biotoflow}

Similar to the SoftCite corpus experiments, we conducted experiments on BioToFlow using NLStruct and divided the data as follows: 75\% of the corpus (39 articles) in the training set, with two-thirds of these (26 articles) used for model training and one-third (13 articles) for validation and 25\% (13 articles) formed the test set. The results obtained are presented in Table \ref{tab_biotoflow}. 

\nprounddigits{1}
\begin{table}[!ht]
\caption{Average performance on five splits and five random seeds (with standard deviation) of percentage scores obtained with SciBERT and NLStruct in relaxed mode on the BioToFlow corpus. \textbf{Overall-focused} results are directly comparable to Table \ref{tab_softcite}.} \label{tab_biotoflow}
\footnotesize
\centering
\resizebox{\textwidth}{!} {
\begin{tabular}{llllllll}
\toprule
\textbf{Entities} & \textbf{P}  & \textbf{R}  & \textbf{F1} & \textbf{Entities}   & \textbf{P}  & \textbf{R} & \textbf{F1} \\
\midrule
Biblio            & \numprint{97,90}{\scriptsize $\pm$\numprint{0,41}}& \numprint{97,42}{\scriptsize $\pm$\numprint{1,00}} & \numprint{97,65}{\scriptsize $\pm$\numprint{0,36}} & LibraryPackage      & \numprint{39,78}{\scriptsize $\pm$\numprint{12,31}}& \numprint{30,93}{\scriptsize $\pm$\numprint{6,74}} & \numprint{33,92}{\scriptsize $\pm$\numprint{7,35}}  \\
Container         & \numprint{78,28}{\scriptsize $\pm$\numprint{11,22}}& \numprint{80,24}{\scriptsize $\pm$\numprint{12,45}} & \numprint{78,87}{\scriptsize $\pm$\numprint{11,75}} & {\scriptsize MgmtSystem}    & \numprint{72,87}{\scriptsize $\pm$\numprint{5,64}}& \numprint{90,85}{\scriptsize $\pm$\numprint{3,99}} & \numprint{80,60}{\scriptsize $\pm$\numprint{3,36}} \\
Data              & \numprint{70,91}{\scriptsize $\pm$\numprint{2,13}}& \numprint{57,09}{\scriptsize $\pm$\numprint{2,29}} & \numprint{63,22}{\scriptsize $\pm$\numprint{1,87}}& Method              & \numprint{38,56}{\scriptsize $\pm$\numprint{2,62}}& \numprint{68,27}{\scriptsize $\pm$\numprint{3,43}} & \numprint{49,21}{\scriptsize $\pm$\numprint{2,92}}  \\
Database          & \numprint{63,82}{\scriptsize $\pm$\numprint{3,62}}& \numprint{54,77}{\scriptsize $\pm$\numprint{4,28}} & \numprint{58,69}{\scriptsize $\pm$\numprint{1,84}} & Parameter           & \numprint{53,16}{\scriptsize $\pm$\numprint{3,26}}& \numprint{53,05}{\scriptsize $\pm$\numprint{4,82}} & \numprint{52,89}{\scriptsize $\pm$\numprint{2,91}}  \\
Description       & \numprint{65,36}{\scriptsize $\pm$\numprint{3,94}}& \numprint{59,27}{\scriptsize $\pm$\numprint{1,79}} & \numprint{62,05}{\scriptsize $\pm$\numprint{1,47}} & {\scriptsize ProgLanguage} & \numprint{86,86}{\scriptsize $\pm$\numprint{2,12}}& \numprint{87,22}{\scriptsize $\pm$\numprint{2,87}}& \numprint{86,94}{\scriptsize $\pm$\numprint{2,38}}  \\
Environment       & \numprint{50,26}{\scriptsize $\pm$\numprint{17,94}}& \numprint{79,36}{\scriptsize $\pm$\numprint{10,09}} & \numprint{60,29}{\scriptsize $\pm$\numprint{17,31}} & Tool                & \numprint{77,71}{\scriptsize $\pm$\numprint{3,20}}& \numprint{72,18}{\scriptsize $\pm$\numprint{2,83}} & \numprint{74,75}{\scriptsize $\pm$\numprint{1,04}}  \\
File              & \numprint{84,28}{\scriptsize $\pm$\numprint{0,53}}& \numprint{82,48}{\scriptsize $\pm$\numprint{1,44}} & \numprint{83,35}{\scriptsize $\pm$\numprint{0,64}} & Version             & \numprint{86,84}{\scriptsize $\pm$\numprint{2,60}}& \numprint{90,80}{\scriptsize $\pm$\numprint{2,01}} & \numprint{88,72}{\scriptsize $\pm$\numprint{0,86}}  \\
Hardware          & \numprint{65,76}{\scriptsize $\pm$\numprint{1,18}}& \numprint{79,16}{\scriptsize $\pm$\numprint{4,42}} & \numprint{71,72}{\scriptsize $\pm$\numprint{1,70}} & WorkflowName        & \numprint{67,10}{\scriptsize $\pm$\numprint{2,18}}& \numprint{77,82}{\scriptsize $\pm$\numprint{4,39}} & \numprint{71,96}{\scriptsize $\pm$\numprint{2,09}}  \\
\midrule
\textbf{Overall}           & \numprint{71,20}{\scriptsize $\pm$\numprint{1,32}}& \numprint{69,62}{\scriptsize $\pm$\numprint{0,90}} & \numprint{70,39}{\scriptsize $\pm$\numprint{0,84}} & 
\textbf{Overall-focused}   & \numprint{85,54}{\scriptsize $\pm$\numprint{1,19}} & \numprint{83,37}{\scriptsize $\pm$\numprint{1,16}} & \numprint{84,42}{\scriptsize $\pm$\numprint{0,47}}      \\     
\bottomrule
\end{tabular}
}
\end{table}
\nprounddigits{0}

The first results show that training a model on our little corpus obtains higher results for all the entities (see Tables \ref{tab_llm_new}, \ref{tab_softcite}, and \ref{tab_biotoflow}).

\subsection{Fusion of \textit{SoftCite} and \textit{BioToFlow} Corpora}

The previous results show that SoftCite and BioToFlow, when used independently, achieve good scores (above 70\%). We also merged the two corpora and trained models on the combined dataset to see the contribution of the SoftCite corpus. Two configurations of the SoftCite corpus were considered: (a) entities were only converted to match BioToFlow entity names and (b) each SoftCite article was annotated using predictions from a model trained on BioToFlow.

The scores for specific entities are provided in Table \ref{tab_soft_fusion}. The results suggest that the SoftCite corpus, in its silver form, achieves a slightly higher F1 score compared to a model trained only on BioToFlow (Table \ref{tab_biotoflow}). These corpus fusions resulted in some entity scores increasing while others decreased.  

\nprounddigits{1}
    \begin{table}[!ht]
    \caption{Average performance metrics on five splits and five random seeds (with standard deviation) of percentage scores obtained using SciBERT and NLStruct in relaxed mode on the fusion of BioToFlow and SoftCite corpora.} \label{tab_soft_fusion}
    \footnotesize
    \setlength{\tabcolsep}{6pt}
    \centering
    \begin{tabular}{llllllll}
        \toprule
                            & \multicolumn{3}{c}{\textbf{With SoftCite conversion}}                                             & \multicolumn{3}{c}{\textbf{With SoftCite silver}}                                                 \\
                             \cmidrule(lr){2-4} \cmidrule(lr){5-7}
                            & \multicolumn{1}{c}{\textbf{P}} & \multicolumn{1}{c}{\textbf{R}} & \multicolumn{1}{c}{\textbf{F1}} & \multicolumn{1}{c}{\textbf{P}} & \multicolumn{1}{c}{\textbf{R}} & \multicolumn{1}{c}{\textbf{F1}} \\
        \midrule
        Data                & \numprint{56,89}{\scriptsize $\pm$\numprint{2,25}} & \numprint{64,08}{\scriptsize $\pm$\numprint{1,52}}  & \numprint{60,12}{\scriptsize $\pm$\numprint{0,72}}  & \numprint{75,83}{\scriptsize $\pm$\numprint{2,60}}  & \numprint{56,81}{\scriptsize $\pm$\numprint{2,11}} & \numprint{64,89}{\scriptsize $\pm$\numprint{0,80}}                      \\
        Description         & \numprint{62,90}{\scriptsize $\pm$\numprint{3,65}} & \numprint{62,49}{\scriptsize $\pm$\numprint{1,33}}  & \numprint{62,63}{\scriptsize $\pm$\numprint{1,17}}  & \numprint{70,98}{\scriptsize $\pm$\numprint{0,83}}  & \numprint{57,43}{\scriptsize $\pm$\numprint{0,92}} & \numprint{63,47}{\scriptsize $\pm$\numprint{0,71}}                      \\
        LibraryPackage      & \numprint{36,22}{\scriptsize $\pm$\numprint{7,86}} & \numprint{38,09}{\scriptsize $\pm$\numprint{6,25}}  & \numprint{36,06}{\scriptsize $\pm$\numprint{6,96}}  & \numprint{25,33}{\scriptsize $\pm$\numprint{6,35}}  & \numprint{37,23}{\scriptsize $\pm$\numprint{8,91}} & \numprint{28,95}{\scriptsize $\pm$\numprint{5,71}}                      \\
        ProgLanguage & \numprint{88,57}{\scriptsize $\pm$\numprint{5,42}} & \numprint{87,62}{\scriptsize $\pm$\numprint{3,99}}  & \numprint{87,91}{\scriptsize $\pm$\numprint{2,40}}  & \numprint{90,29}{\scriptsize $\pm$\numprint{1,30}}  & \numprint{88,25}{\scriptsize $\pm$\numprint{2,97}} & \numprint{89,18}{\scriptsize $\pm$\numprint{1,43}}                      \\
        Tool                & \numprint{76,92}{\scriptsize $\pm$\numprint{1,27}} & \numprint{72,24}{\scriptsize $\pm$\numprint{1,16}}  & \numprint{74,45}{\scriptsize $\pm$\numprint{0,92}}  & \numprint{78,84}{\scriptsize $\pm$\numprint{1,87}}  & \numprint{69,97}{\scriptsize $\pm$\numprint{0,93}} & \numprint{74,13}{\scriptsize $\pm$\numprint{1,10}}                      \\ \midrule
        \textbf{Overall}             & \numprint{64,40}{\scriptsize $\pm$\numprint{0,95}} & \numprint{73,57}{\scriptsize $\pm$\numprint{0,89}}  & \numprint{68,66}{\scriptsize $\pm$\numprint{0,27}}  & \numprint{73,00}{\scriptsize $\pm$\numprint{0,47}}  & \numprint{68,52}{\scriptsize $\pm$\numprint{1,40}} & \numprint{70,68}{\scriptsize $\pm$\numprint{0,55}}     \\ \bottomrule                
        \end{tabular}
        \end{table}
        \nprounddigits{0}

\section{Integration of Knowledge into Language Models}

Since the various tested approaches did not show any significant improvement in NER scores compared to a masked language model trained on BioToFlow, we explored one final approach: the integration of knowledge into language models.

\subsubsection{Used of a Pre-initialized Encoder} In the spirit of work exploiting pretraining for NER \cite{liu-arxiv-21}, we trained models on BioToFlow following the same methodology as in Section~\ref{sec:ner_biotoflow} but with an encoder initialized with one of the models trained during the NER experiments on the SoftCite corpus (see Section~\ref{section_softcite}). The scores of Table~\ref{tab_soft_encodeur} for a subset of entities show that the  results obtained are nearly identical to those achieved with a model whose encoder was not pre-initialized.

\nprounddigits{1}
    \begin{table}
    \caption{Average performance metrics on five splits and five random seeds (with standard deviation) of percentage scores for a subset of entities obtained using SciBERT and NLStruct in relaxed mode on BioToFlow corpus with the encoder initialization.} \label{tab_soft_encodeur}
    \footnotesize
    \centering
    \begin{tabular}{llllllll}
        \toprule
                         & \multicolumn{1}{c}{\textbf{P}} & \multicolumn{1}{c}{\textbf{R}} & \multicolumn{1}{c}{\textbf{F1}} & \multicolumn{1}{c}{\textbf{}} & \multicolumn{1}{c}{\textbf{P}} & \multicolumn{1}{c}{\textbf{R}} & \multicolumn{1}{c}{\textbf{F1}} \\
        \midrule
        Database         & \numprint{63,88}{\scriptsize $\pm$\numprint{3,19}}   & \numprint{56,23}{\scriptsize $\pm$\numprint{3,26}}    & \numprint{59,65}{\scriptsize $\pm$\numprint{1,26}}     & Method              & \numprint{38,15}{\scriptsize $\pm$\numprint{1,21}}    & \numprint{68,13}{\scriptsize $\pm$\numprint{3,27}}   & \numprint{48,86}{\scriptsize $\pm$\numprint{1,58}}                      \\
        Environment      & \numprint{47,65}{\scriptsize $\pm$\numprint{18,89}}   & \numprint{79,99}{\scriptsize $\pm$\numprint{14,33}}   & \numprint{58,53}{\scriptsize $\pm$\numprint{20,14}}    & {\scriptsize ProgLanguage}        & \numprint{87,14}{\scriptsize $\pm$\numprint{1,01}}    & \numprint{85,87}{\scriptsize $\pm$\numprint{5,04}}   & \numprint{86,24}{\scriptsize $\pm$\numprint{2,53}}                      \\
        LibraryPackage   & \numprint{40,22}{\scriptsize $\pm$\numprint{11,48}}  & \numprint{32,38}{\scriptsize $\pm$\numprint{5,64}}    & \numprint{35,08}{\scriptsize $\pm$\numprint{5,92}}     & Tool                & \numprint{77,29}{\scriptsize $\pm$\numprint{3,11}}    & \numprint{72,07}{\scriptsize $\pm$\numprint{2,19}}   & \numprint{74,50}{\scriptsize $\pm$\numprint{0,55}}                      \\
        MgmtSystem & \numprint{72,73}{\scriptsize $\pm$\numprint{5,85}}    & \numprint{94,38}{\scriptsize $\pm$\numprint{5,65}}    & \numprint{81,95}{\scriptsize $\pm$\numprint{4,17}}     & \textbf{Overall}             & \numprint{71,25}{\scriptsize $\pm$\numprint{0,81}}    & \numprint{69,64}{\scriptsize $\pm$\numprint{1,29}}   & \numprint{70,42}{\scriptsize $\pm$\numprint{0,77}} \\ \bottomrule                     
        \end{tabular}
    \end{table}
    \nprounddigits{0}

\subsubsection{Adding Vocabulary}\textit{Tools} are considered a key entity in our work because each tool has a specific function in the workflow, and it is essential for the overall understanding of the workflow task that all tools are extracted properly. To improve the extraction of bioinformatics tool names, another strategy involves incorporating external knowledge into the encoder. Since the knowledge to inject in our case is made of lists of names, we chose to perform this injection at the level of its vocabulary, similarly to work such as \cite{gee-etal-2022-fast,hong-avocado-21}. Various knowledge bases of bioinformatics tools exist, from which tool and binary names (the executable files or commands used to run the tools in a program) can be extracted. We utilized four knowledge bases: \textit{Biotools} \cite{ison_tools_2016}, \textit{Bioconda} \cite{gruning_bioconda_2018}, \textit{Biocontainers}  \cite{da_veiga_leprevost_biocontainers_2017} and \textit{Bioweb}\footnote{\url{https://bioweb.pasteur.fr/welcome}}. From these knowledge bases, we aggregated the names of tools and binaries to create an enriched vocabulary. This new vocabulary was then injected into the SciBERT language model to improve its ability to recognize bioinformatics tool names. We evaluated this strategy by conducting two experiments on BioToFlow, testing two approaches: (a) fine-tuning the SciBERT model with the new vocabulary and (b) using the enriched vocabulary without additional fine-tuning. In both cases, the embeddings of the added tokens were initialized by averaging the embeddings of their subtokens.

\nprounddigits{1}
Adding bioinformatics tool vocabulary appears to improve the extraction of tool names, particularly when fine-tuning the SciBERT model after incorporating the new vocabulary, with an F1 score increasing from \numprint{74.8}{\scriptsize$\pm$\numprint{1.0}}\% to \numprint{77.0}{\scriptsize$\pm$\numprint{2.8}}\%. As observed in previous experiments, some entities benefit from the added knowledge, while others experience slight decreases in performance (see Table \ref{tab_add_voc}).
Specifically, the \textit{Data} entity shows improvements with vocabulary addition in both fine-tuning and non-fine-tuning scenarios. Entities like \textit{LibraryPackage} and \textit{Tool} also benefit from fine-tuning with the added vocabulary. Conversely, scores for \textit{ManagementSystem} and \textit{ProgrammingLanguage} entities show slight declines.
\nprounddigits{0}

\nprounddigits{1}
        \begin{table}[!ht]
        \caption{Average performance metrics on five splits and five random seeds (with standard deviation) of percentage scores obtained using SciBERT and NLStruct in relaxed mode with the addition of bioinformatics tool vocabulary, with or without fine-tuning.} \label{tab_add_voc}
        \footnotesize
        \setlength{\tabcolsep}{3pt}
        \centering
        \begin{tabular}{llllllll}
            \toprule
                                & \multicolumn{3}{c}{\textbf{Add voc – finetune}} & \multicolumn{3}{c}{\textbf{Add voc – no finetune}} \\
                \cmidrule(lr){2-4} \cmidrule(lr){5-7}
                                & \textbf{P}     & \textbf{R}     & \textbf{F1}   & \textbf{P}      & \textbf{R}     & \textbf{F1}     \\
            \midrule
            Data                & \numprint{69,98}{\scriptsize $\pm$\numprint{7,20}} & \numprint{63,42}{\scriptsize $\pm$\numprint{10,87}}    & \numprint{66,36}{\scriptsize $\pm$\numprint{8,86}}   & \numprint{71,66}{\scriptsize $\pm$\numprint{2,35}}      & \numprint{56,94}{\scriptsize $\pm$\numprint{2,82}}     & \numprint{63,41}{\scriptsize $\pm$\numprint{2,27}}      \\
            LibraryPackage      & \numprint{52,52}{\scriptsize $\pm$\numprint{7,12}} & \numprint{57,43}{\scriptsize $\pm$\numprint{15,55}}    & \numprint{52,26}{\scriptsize $\pm$\numprint{10,00}}   & \numprint{37,11}{\scriptsize $\pm$\numprint{8,00}}      & \numprint{30,38}{\scriptsize $\pm$\numprint{7,21}}     & \numprint{32,73}{\scriptsize $\pm$\numprint{6,02}}      \\
            ManagementSystem    & \numprint{75,60}{\scriptsize $\pm$\numprint{7,88}} & \numprint{76,84}{\scriptsize $\pm$\numprint{6,70}}    & \numprint{75,54}{\scriptsize $\pm$\numprint{7,03}}   & \numprint{69,73}{\scriptsize $\pm$\numprint{3,87}}      & \numprint{93,42}{\scriptsize $\pm$\numprint{3,33}}     & \numprint{79,72}{\scriptsize $\pm$\numprint{1,73}}      \\
            ProgrammingLanguage & \numprint{81,93}{\scriptsize $\pm$\numprint{12,58}} & \numprint{82,09}{\scriptsize $\pm$\numprint{11,49}}    & \numprint{81,90}{\scriptsize $\pm$\numprint{11,96}}   & \numprint{86,00}{\scriptsize $\pm$\numprint{2,06}}      & \numprint{85,79}{\scriptsize $\pm$\numprint{2,33}}     & \numprint{85,83}{\scriptsize $\pm$\numprint{2,11}}      \\
            Tool                & \numprint{77,23}{\scriptsize $\pm$\numprint{3,70}} & \numprint{76,90}{\scriptsize $\pm$\numprint{2,71}}    & \numprint{77,01}{\scriptsize $\pm$\numprint{2,79}}   & \numprint{74,49}{\scriptsize $\pm$\numprint{2,49}}      & \numprint{73,92}{\scriptsize $\pm$\numprint{2,62}}     & \numprint{74,12}{\scriptsize $\pm$\numprint{0,33}}      \\
            \midrule
            \textbf{Overall}             & \numprint{70,43}{\scriptsize $\pm$\numprint{1,69}} & \numprint{70,21}{\scriptsize $\pm$\numprint{1,52}}    & \numprint{70,27}{\scriptsize $\pm$\numprint{0,43}}   & \numprint{70,60}{\scriptsize $\pm$\numprint{1,23}}      & \numprint{69,53}{\scriptsize $\pm$\numprint{1,23}}     & \numprint{70,05}{\scriptsize $\pm$\numprint{0,86}}   \\ \bottomrule  
            \end{tabular}
            \end{table}
\nprounddigits{0}

\section{Conclusion}

Based on a workflow representation schema, we developed a small entity-rich corpus, BioToFlow. Using this corpus, we experimented with various entity extraction techniques, including decoder-based models, which achieved scores below 50\%, and encoder-based methods. These included leveraging a large corpus with annotations covering a subset of the schema and our BioToFlow corpus. A final approach sought to incorporate additional knowledge into the models. The most effective strategy involved training a model on our small corpus while integrating a vocabulary list specific to bioinformatics tools, which improved the extraction of this entity type. All these experiments and combinations of approaches demonstrate that bioinformatics workflows represent a highly specialized branch of biology. This implies that named entity extraction requires a large amount of domain-specific annotated data, as well as the use of language models tailored to bioinformatics. To improve entity extraction, additional few-shot learning approaches should be explored, such as using NER models trained to identify any type of entity, like GLiNER \cite{zaratiana-etal-2024-gliner}. 
To further extend the approach of incorporating domain knowledge, we will also investigate various ways to enhance the embeddings of the added vocabulary, starting with their adaptation during the training of the NER models, similarly to \cite{hong-avocado-21}. More globally, the entity linking between the components extracted from workflows described in scientific literature and those found in code from public repositories is an essential task that we will consider to merge information from both sources.

\begin{credits}
\subsubsection{\ackname} We sincerely thank N. Bossut, A. Gaignard, G. Marchment, M. Schmit, H. Menager, F. Lemoine for their help with the annotation process. We also thank M. Naguib for his help with the decoder-based experiments. This work has received support from the French government (Agence Nationale pour la Recherche) under the France 2030 program grant agreement ANR-22-PESN-0007 (ShareFAIR).

\subsubsection{\discintname}
The authors have no competing interests. 
\end{credits}

\appendix

\section{Hyperparameters used with NLStruct}\label{nlstruct_h}
\begin{table}[!ht]
\caption{Hyperparameters used with NLStruct.} \label{hyperparams_nlstruct}
\centering
\setlength{\tabcolsep}{3pt}
\begin{tabular}{llll}
\toprule
\textbf{Parameter} & \textbf{Value} & \textbf{Parameter} & \textbf{Value} \\ \midrule
Base encoder  & \begin{tabular}[c]{@{}l@{}}BERT-base-uncased\\  SciBERT\_scivocab\_uncased\end{tabular} & Learning rate  & 1e-3 \\
Sequence length    & 256 & Dropout   & \numprint{0,1} \\
Train iteration    & \numprint{4000} & Warmup ratio       & \numprint{0,1} \\
Optimizer   & AdamW    & Random seed        & 1 - 8 - 22 - 42 -100 \\  \bottomrule          
\end{tabular}
\end{table}

\end{document}